\documentclass[3p,times,procedia]{elsarticle}

\usepackage{ecrc}

\volume{00}

\firstpage{1}

\runauth{}

\jid{procs}

\CopyrightLine{2011}{Published by Elsevier Ltd.}

\usepackage{amssymb}
\usepackage{xcolor}  
\usepackage{float}
\usepackage{amsmath}
\usepackage{multirow}
\usepackage{booktabs}
\usepackage{booktabs}
\usepackage[textsize=scriptsize]{todonotes}
\usepackage{CJKutf8}
\usepackage{cite}
\usepackage{amsmath,amssymb,amsfonts}
\usepackage{algorithmic}
\usepackage{graphicx}
\usepackage{textcomp}
\usepackage{xcolor}
\usepackage{amsmath}
\usepackage{subfigure} 

\usepackage[figuresright]{rotating}
\usepackage{multirow}
\usepackage{booktabs}

\begin{document}

\begin{frontmatter}



\dochead{}

\title{Cross-Domain Car Detection Model with Integrated Convolutional Block Attention Mechanism}


\author[author1]{$\rm Haoxuan\,\,Xu^{**}$}
\ead{202020120237@mail.sdu.edu.cn}
\author[author1]{$\rm Songning\,\, Lai^{**}$}
\ead{sonly@mail.sdu.edu.cn}
\author[author2]{$\rm Xianyang\,\, Li$}
\ead{lxy6688@stu.ouc.edu.cn}
\author[author1]{$\rm Yang\,\, Yang^{*}$}
\ead{yyang@sdu.edu.cn}

\cortext[cor1]{Corresponding author}
\cortext[2]{These authors contributed equally}
\address[author1]{School of Information Science and Engineering, Shandong University, Qingdao, China}
\address[author2]{School of Economics, Ocean University of China, Qingdao, China}

\begin{abstract}
Car detection, particularly through camera vision, has become a major focus in the field of computer vision and has gained widespread adoption. While current car detection systems are capable of good detection, reliable detection can still be challenging due to factors such as proximity between the car, light intensity, and environmental visibility. To address these issues, we propose cross-domain \textbf{C}ar \textbf{D}etection \textbf{M}odel with integrated convolutional block \textbf{A}ttention mechanism(CDMA) that we apply to car recognition for autonomous driving and other areas. CDMA includes several novelties: 1)Building a complete cross-domain target detection framework. 2)Developing an unpaired target domain picture generation module with an integrated convolutional attention mechanism which specifically emphasizes the car headlights feature. 3)Adopting Generalized Intersection over Union (GIOU) as the loss function of the target detection framework. 4)Designing an object detection model integrated with two-headed Convolutional Block Attention Module(CBAM). 5)Utilizing an effective data enhancement method. To evaluate the model's effectiveness, we performed a reduced will resolution process on the data in the SSLAD dataset and used it as the benchmark dataset for our task. Experimental results show that the performance of the cross-domain car target detection model improves by 40\% over the model without our framework, and our improvements have a significant impact on cross-domain car recognition,exceeding most advaenced crossdomain models.

\end{abstract}


\end{frontmatter}


\section{Introduction}

Deep learning techniques enable the emergence of state-of-the-art models for solving target detection tasks\citep{Ren2015faster,redmon2016yolo}. However, these techniques rely heavily on data and the accuracy of the training dataset. Current deep learning tasks mostly assume similar data distributions for both the training and test sets, making the task substantially less challenging. Ideally, object detectors should maintain high performance in detecting a given object despite significant variability in the data distribution. For example, the model should perform well on test data with different environments and angles from the training data. When the data distribution changes, we refer to this as a change in domain, which can lead to decreased object detection accuracy. Researchers in the field of domain adaptation have defined and explored this problem, proposing the task of learning the key features of an object from the source domain.

Data annotation is a costly process for target detection, making it crucial to develop a model that can effectively detect targets in the target domain. It also presents a challenging goal for Unsupervised Domain Adaptation (UDA)\citep{wang2018deep,Tzeng2017adversarial,ganin2015unsupervised}, where only unlabelled target data is available in addition to labelled source data. Moreover, the training data may have been collected under varying scenario conditions, which is known as multi-source domain adaptation.

The primary approach in UDA is to learn domain invariant feature representations by aligning the source and target domains. In target detection, recent research has focused on how to learn aligned features and how to induce feature alignment. For instance, \citep{chen2018domain,he2019multi} propose adapting the backbone network and all internal image elevation features, as well as auxiliary features extracted from object suggestions using adversarial training\citep{ganin2015unsupervised}. \citep{xu2020cross} argues for the advantages of aggregating object suggestions before alignment and suggests compressing all suggestions into a single class of prototype vectors prior to alignment using contrast loss induction.

Despite the success of many CNN-based target detection methods driven by deep convolutional networks (CNNs) \citep{girshick2014rich,Ren2015faster,zhu2020deformable}, cross-domain detection remains a highly desirable and challenging task. Given the difficulty of obtaining labels, a method that can transform images from one domain to another can enable cross-domain label transformation. Generative Adversarial Networks (GANs) \citep{mirza2014conditional,isola2017image,wang2018high,zhu2017unpaired}have emerged as a powerful tool for constructing image generation methods that address the problem of image transformation. In recent years, GAN-based image transformation methods have shown advanced image translation between different domains in an unsupervised manner. For instance, \citep{zhu2017unpaired}use supervised techniques to build frameworks to propose CycleGAN, which can transform two types of images in both directions without pairwise matching. StarGAN\citep{choi2018stargan} and UNIT\citep{khandelwal2021unit} can transform images in multiple domain styles simultaneously.

Car detection faces significant challenges due to the differences in lighting conditions between day and night. Current research mainly focuses on improving the robustness of models to adapt to various lighting conditions. However, existing methods often fail to fully consider the importance of car headlights in car detection. Therefore, CDMA uses an attention mechanism module to emphasize the importance of car headlights in the image transfer process, thereby improving the accuracy of the model in day and night image detection. Our research fills the gap in current research and achieves good results in experiments.

In this study, we use the detection of car as a test case for our improved cross-domain object detection method. Our approach incorporates the concept of generative adversarial image generation and combines attention mechanisms from the latent space to exploit different features in the image, resulting in improved image generation for adversarial generative training. This results in better generators and decoders for GAN networks, allowing for better achievement of target-domain style image generation goals. Additionally, we enhance the detection efficiency of the car detector in the target domain by convolving the attention mechanism in both space and channel, using generalized cross union to increase the effectiveness of model correction.

In summary, our key contributions are as follows:

1)A comprehensive framework for is proposed for nighttime target detection. By fine-tuning the target detector using the image generator in the target domain based on source domain training, CDMA achieves high-accuracy cross-domain target detection in the absence of the target domain, improving mAP by 18.55\% over original target detection results.

2)A image generator is created which learns the primary features of blackout vehicles through an unpair target domain image generation module with an integrated convolutional attention mechanism.

3)The overall detection performance of the Faster R-CNN model is enhanced by using Generalized Intersection over Union as the loss function of the target detection framework, combined with two-headed Convolutional Block Attention Module,which we name it Attention R-CNN.

4)An effective data augmentation method is employed to expand the dataset effectively, even with a limited training set.

Overall, these contributions demonstrate that our approach can effectively address the challenges of cross-domain target detection and significantly improve detection accuracy. We believe that our work will provide valuable insights for researchers and practitioners in this field.

\section{Related Work}
\textbf{Object detection:} Classical object detection relies on manually extracted features and sliding window traversal of image pyramids for classification\citep{girshick2014rich}. However, deep convolutional networks trained on large-scale data are gaining popularity. These deep learning algorithms can be classified into two types: frame-based \citep{yu2016unitbox,zhang2018generalized}and frame-free\citep{law2018cornernet}. Frame-based algorithms can be further divided into one-stage\citep{redmon2016yolo,tan2020efficientdet} and two-stage \citep{Ren2015faster,cao2020maskcnn}detection algorithms, while frame-free algorithms can be divided into keypoint and centroid methods. Faster R-CNN is widely used due to its good performance and open implementation. It utilizes a region proposal network (RPN) to generate candidate bounding boxes in the first stage, and in the second stage, it employs RoIPool and a fully connected layer to extract features for each candidate box. RPN and Fast R-CNN networks are trained jointly in an end-to-end manner using a multi-task loss function that combines the classification loss and regression loss.  The classification loss measures the accuracy of the predicted class labels, while the regression loss measures the accuracy of the predicted bounding box offsets. By improving Faster R-CNN, we aim to enhance our detection model as well.

\textbf{Unsupervised Cross-Domain Object Detection:} Unsupervised cross-domain object detection involves detecting objects in an unsupervised target domain using only labeled data from the source domain\citep{inoue2018cross,chen2018dafaster,yu2019Unsupervised,wang2018unsupervised}. In such tasks, semi-supervised learning-based methods utilize semi-supervised learning. This involves training a base model with labeled data from the source domain and an object detection model. The model is then used to predict the unlabeled data in the target domain, and the predicted pseudo-labeled data and labeled data from the source domain are combined as the training data for semi-supervised learning \citep{inoue2018cross,yu2019unspervised}. Adversarial learning-based methods establish an adversarial network between the source and target domains, enabling cross-domain target detection by minimizing the distance between the source and target domains while maximizing the accuracy of the domain classifier\citep{chen2018dafaster,Tzeng2017adversarial,Vu2019advent}.

\textbf{Dark Scene Image Generation:} By simulating the lighting conditions in low-light environments, we can generate high-quality dark scene images.Early methods for generating dark scene images were primarily based on statistical models, with the Gaussian Mixture Model [GMM] being used to model the transformation between night and day scenes\citep{nurhadiyatna2013backgroud}. More recent methods employ physical models to simulate the propagation process of light in dark scenes by establishing a light transmission model, resulting in high-quality dark scene images. Physical models are preferred, as they better simulate real-world lighting conditions and have been widely used in dark scene image generation tasks. Deep learning-based physical models, such as DeepISP\citep{schwartz2018deepisp} and DeepUPE\citep{he2018deep}, have also been developed and achieved promising results in generating dark scene images.
GAN-based methods are currently among the most widely used methods for generating dark scene images. These methods train the generator and discriminator to produce images that closely resemble real dark scene images, while ensuring that the discriminator can distinguish between generated and real images with high accuracy. Models such as DCGAN\citep{radford2015unsupervised}, Pix2Pix\citep{isola2017image}, and CycleGAN\citep{zhu2017unpaired} have achieved excellent results in dark scene image generation tasks. CDMA is mainly based on CycleGAN.

\textbf{Attention Mechanism:} The attention mechanism is a technique that simulates the human visual attention mechanism, allowing the model to focus more on critical information in the input sequence and improve overall performance. In recent years, the attention mechanism has been widely adopted in computer vision for various tasks, including image classification, object detection, and image generation. For example, Xu et al\citep{xu2015show}. proposed a Convolutional Neural Network that utilizes attention mechanism to emphasize significant regions in the input image and improve classification accuracy. Similarly, Chen et al\citep{chen2017deeplab}. developed an object detection model based on attention mechanism that can automatically prioritize the target area and enhance detection accuracy. The Convolutional Block Attention Module (CBAM) \citep{woo2018cbam}is a deep learning model that leverages attention mechanism to boost performance and accuracy. By integrating attention mechanism into a convolutional neural network, CBAM can automatically learn the importance of different regions in an image and better extract image features, resulting in improved classification and detection accuracy.

\section{Methods}
In this section, we present our main model that utilizes image transformation and advanced object detectors to achieve cross-domain car detection. We begin with a problem formulation and then outline the key components of CDMA, which showcases its capability to generalize across day-night domains and different constituent structures. We leverage our brand new model to combine the best performing components through a training strategy designed for cross-domain object detection.

\textbf{Problem formulation:} In unsupervised domain adaptation for object detection, we start with the source domain consisting of $N_s$ labeled images, $S=\left\{\left(x_i^S, y_i^S, B_i^S\right)\right\}_{i=1}^{N_S}$, where $x_i^S$ represents the image, $y_i^S$ denotes the type of image label box, and $B_i^S$ indicates the location of the label box. On the other hand, for the test set of $N_t$ target domains, $T=\left\{x_i^{T}\right\}_{i=1}^{N_T}$, we only have the images themselves without any labels. The primary objective of the UDA method is to train object detectors that can perform well on the target domain, despite the different characteristics of the images between the two domains.

\begin{figure}[ht]
\centering
\includegraphics[width=1\linewidth]{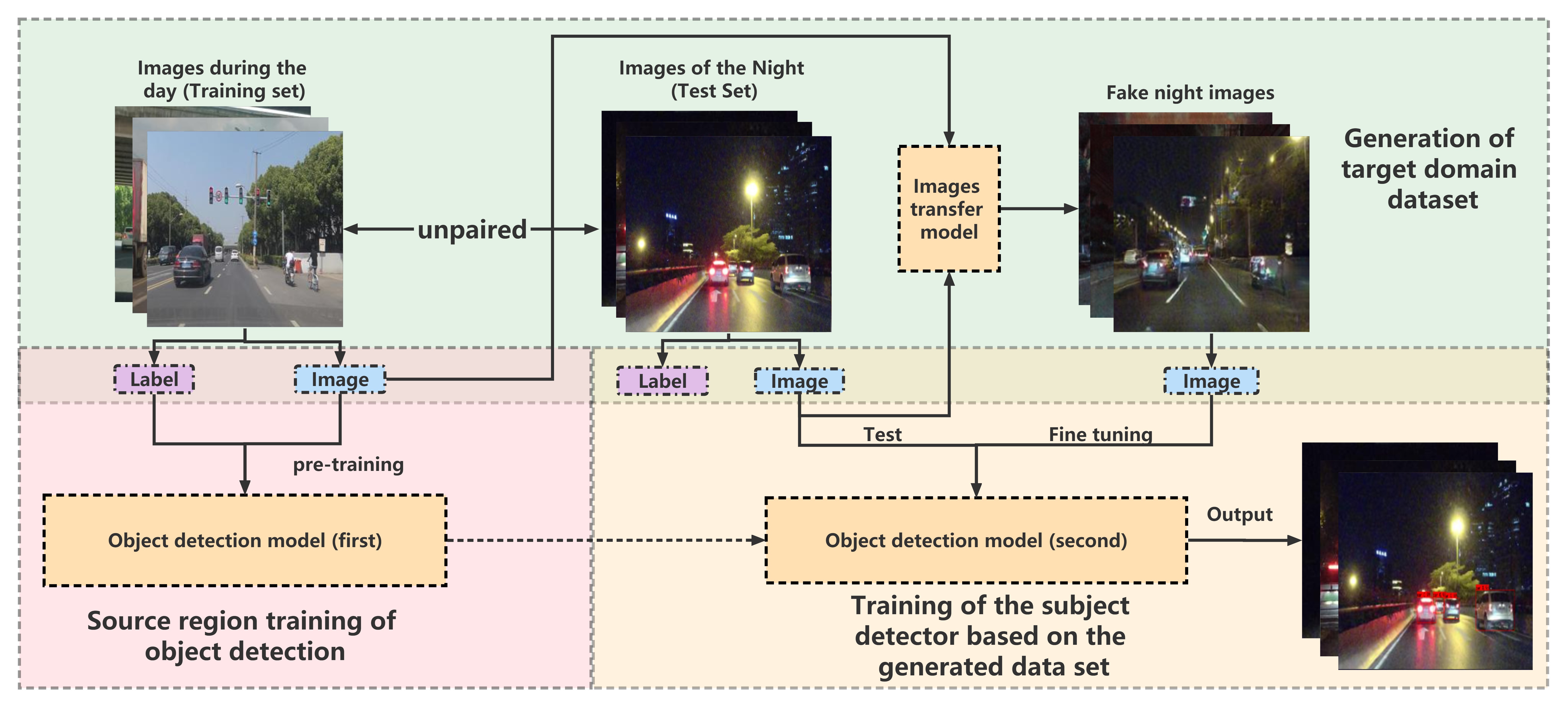}
\caption{\label{fig:overview}
Illustration of Cross-Domain car Object Detection Model:By utilizing source domain images to learn daytime car features within the source domain, the proposed method generates images with dark scene characteristics through the image generator and fine-tunes a target detector to effectively identify cross-domain vehicles.}
\end{figure}

\subsection{Overview}
Our main model consists of three main parts: (i) source domain training for object detection, (ii) generation of the target domain dataset, and (iii) training of the object detector on the generated dataset. As shown in Fig ~\ref{fig:overview}, we first learn and train the source domain dataset using an improved Faster R-CNN model (as mentioned in 3.2) to capture the characteristic information of vehicles during the day. Next, we use an unsupervised image-to-image style converter to generate nighttime target domain images based on the daytime source domain images. Since the converter only modifies the style and appearance of the image, the position of the original car in the image remains unchanged. Hence, we continue to use the labels from the original dataset for the generated images. Finally, we perform fine-tuning on the trained object detector using the generated dark target domain images to learn the features of the target domain and achieve cross-domain object detection.

\subsection{Source Domain Training for Object detection}
\begin{figure}[ht]
\centering
\includegraphics[width=1\linewidth]{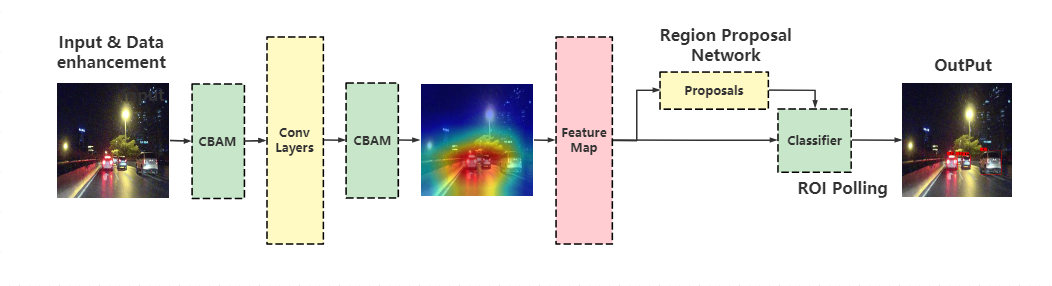}
\caption{\label{fig:object dection}
The object detection model follows a specific process.Our Faster-CNN model incorporates the CBAM attention mechanism before and after the convolutional layer.}
\end{figure}

Car detection utilizes a generic object detector to locate cars in an image. Object detectors take a single image as input and output bounding boxes. Typically, object detection is trained using manually labeled bounding boxes and corresponding images. However, since our objective is to perform unsupervised car detection in dark scenes, we do not have access to corresponding manual annotations. Therefore, we use style-transformed images from the source domain (daytime) for training, as discussed in (3.3). In our work, we adopt Faster R-CNN\citep{Ren2015faster} as the basic framework for optimization. While other models are available\citep{redmon2016yolo,tan2020efficientdet,cao2020maskcnn}, we choose this model due to its openness and excellent performance, CDMA for object dection show as Fig ~\ref{fig:object dection},we incorporates the CBAM attention mechanism before and after the convolutional layer, and employs the GIOU loss function during model training.

Faster R-CNN is a fundamental algorithm for object detection that comprises of two modules. The first network proposes regions that potentially contain objects, while the second network classifies each proposed region and performs bounding box regression. The trained Fast R-CNN network is then incorporated as part of the RPN network, optimizing the entire Faster R-CNN model in an end-to-end training process.

\textbf{Convolutional Attention Module:} We have integrated a module to improve the performance of convolutional neural networks in the backbone network of Faster RCNN. This module consists of two sub-modules: the channel attention module and the spatial attention module\citep{woo2018cbam}.

The channel attention module conducts global max pooling and global average pooling on each channel output of the convolutional layer. A concatenation operation is then performed based on the channels to obtain the attention weight of each channel through the sigmoid activation of two fully connected layers. The final weight is then multiplied by the original convolution layer output to obtain the enhanced feature map.

The spatial attention module conducts adaptive average pooling and maximum pooling on the feature map of each channel. The two pooling results are then passed through a convolutional layer to obtain the corresponding weights. Finally, these weights are added to obtain the attention weight of each spatial location for weighting the feature map. In our model framework, the CBAM attention mechanism after the first convolutional layer and after the last convolutional layer, respectively, makes the model pay more attention to shallow features and deep features.

\textbf{GIOU LOSS\citep{rezatofighi2019generalized}:}
IoU Loss\citep{jiang2018acquisition} is a frequently used loss function for object detection that assesses the overlap between the detected box and the true box. However, when there is considerable overlap between the detection box and the true box, the IoU Loss becomes small, resulting in an unstable training process. To address this, we have introduced GIOU LOSS in our object detector.

\begin{equation}
\begin{aligned}
IoU & =\frac{|A \cap B|}{|A \cup B|} 
\end{aligned}
\end{equation}

\begin{equation}
\begin{aligned}
G I o U & =I o U-\frac{|C \backslash(A \cup B)|}{|C|}
\end{aligned}
\end{equation}

As shown in Equation 1, A represents the Ground Truth, B represents the prediction box, and C represents the closure of the two regions .In GIoU Loss, the closure is defined as the smallest rectangle parallel to the coordinate axis that surrounds the two rectangular regions.

The IoU function is designed to measure the overlap between the detected and true box, but it has certain limitations that can lead to unstable training. One major limitation is that it does not account for the difference in area and aspect ratio between the two boxes, which can affect its accuracy. The GIOU function addresses this limitation by incorporating the difference in area and aspect ratio between the two boxes, allowing for a more accurate measurement of overlap. It also includes a penalty term for the distance between the center point of the detected box and the true box, resulting in a more precise measure of the difference between the two boxes. GIOU takes into account the area difference between two bounding boxes, so it can measure the similarity between them more accurately. GIOU avoids the case that the denominator is 0 when it is calculated, so it can avoid the case that it cannot be calculated. GIOU considers the center point distance between two bounding boxes when calculating, so it can reflect the similarity change between them more smoothly. Overall, the GIOU function provides a more accurate and stable measure of overlap, leading to improved performance in object detection models. Due to our fine-tuning strategy, the model requires high stability of loss. We believe that GIOU is more suitable for our model framework than other IoUs.

\subsection{Generation of the target domain dataset}
\textbf{Adversarial Training:} CycleGAN\citep{zhu2017unpaired} is a type of generative adversarial network used for image-to-image translation that learns the mapping relationship between different domains without relying on pairwise data. The original CycleGAN consists of two generators and a discriminator. Each generator learns to transform image domains, while each discriminator determines whether the generated images are authentic or not. The generator and discriminator work in opposition to continuously improve the generator's ability to learn image features in the target domain. The loss of CycleGAN is calculated using the adversarial loss of the generator and discriminator, as well as the cycle consistency loss to measure the difference between the input image and the recovered image of two transformations, and the identity loss.
Adversarial Loss: In Equation 3, s and t represent the source and target domains, respectively. The target domain is used to train the generator and discriminator, enabling the generator to produce images that can deceive the discriminator, making it unable to distinguish between real and fake images.
\begin{equation}
\begin{aligned}
L_{\text {gan}}^{s \rightarrow t}=\left({E}_{x \sim X_t}\left[\left(D_t(x)\right)^2\right]+{E}_{x \sim X_s}\left[\left(1-D_t\left(G_{s \rightarrow t}(x)\right)\right)^2\right]\right) 
\end{aligned}
\end{equation}
Cycle Consistency Loss: As shown in Equation 4, this loss is used to ensure that the generator's mapping is bidirectional, meaning that the mapping from "s" to "t" and the mapping from "t" to "s" are inverses of each other. This loss function helps to generate more realistic images.
\begin{equation}
\begin{aligned}
\left.L_{\text {cycle }}^{s \rightarrow t}=\left.{E}_{x \sim X_s}\left[\mid x-G_{t \rightarrow s}\left(G_{s \rightarrow t}(x)\right)\right)\right|_1\right]
\end{aligned}
\end{equation}
Identity Loss: As depicted in Equation 5, this loss guarantees consistency between the input image and the output image. Specifically, the output image processed by the generator should be as similar to the input image as possible.
\begin{equation}
\begin{aligned}
L_{\text{identity}}^{s \rightarrow t}={E}_{x \sim X_t}\left[\left|x-G_{s \rightarrow t}(x)\right|_1\right]
\end{aligned}
\end{equation}
\textbf{Attention Mechanism:} Despite the good results achieved by CycleGAN, the model's performance is not satisfactory when detecting vehicles in the transition from day to night (as shown in the figure). Images in the dark domain often contain less useful information (mostly black pixels), and although CycleGAN can transfer the style of photos to the dark domain, the model's focus does not seem to be on vehicles. As a result, the model fails to effectively extract important features from dark vehicles, which are often interfered with by street lights and headlights.

\begin{figure}[ht]
\centering
\includegraphics[width=0.8\linewidth]{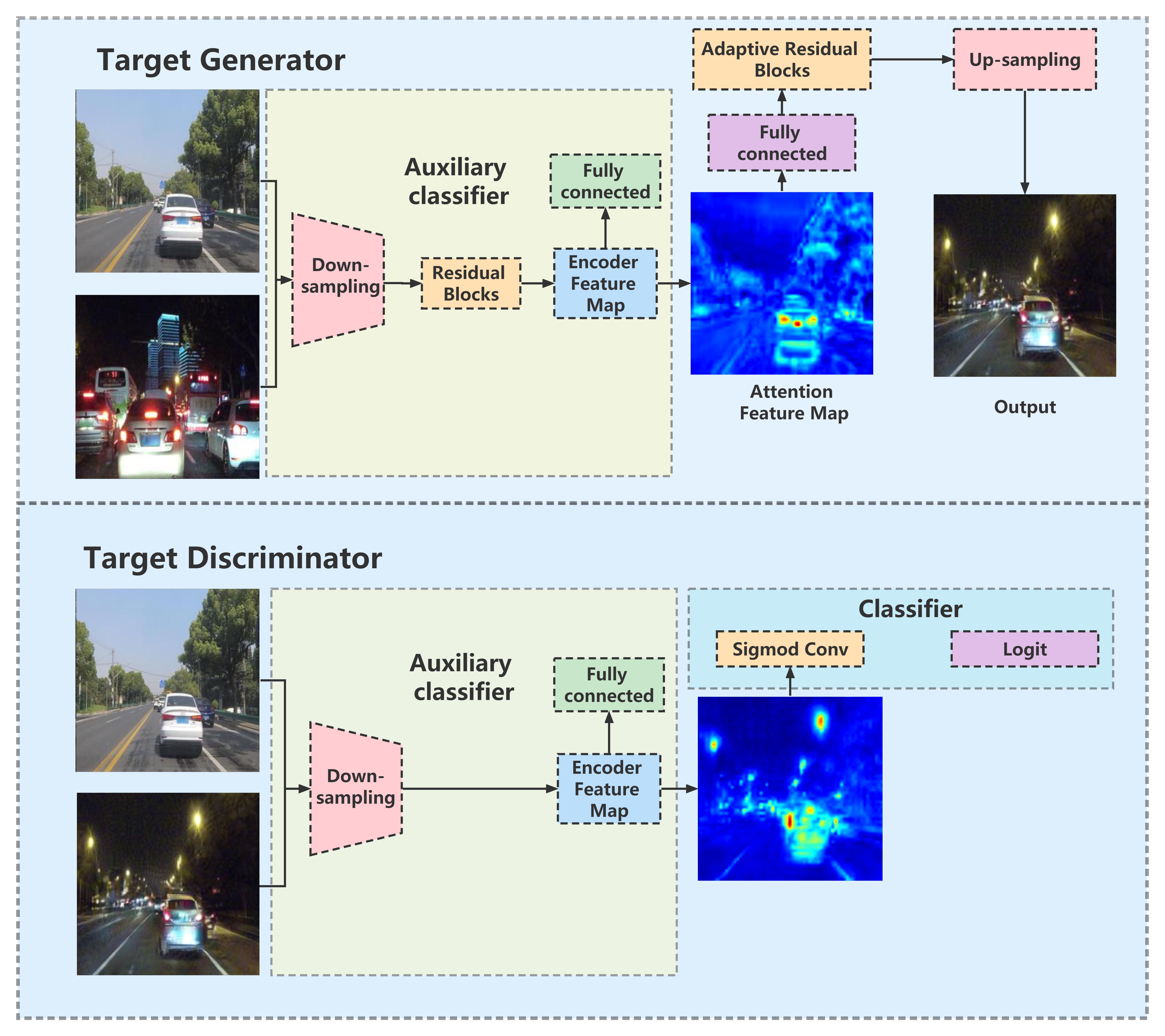}
\caption{\label{fig:cyclegan}The framework of the image generator.We have incorporated an attention mechanism in both the decoder and encoder to generate feature maps that highlight important information. This has led to significant improvements in the image generation quality.}
\end{figure}

To address this issue, we added an appropriate attention mechanism module to the generator and discriminator. Inspired by \citep{kim2019ugat}, we implemented a Class Activation Map (CAM)\citep{zhou2016learning,soydaner2022attention} attention mechanism in CDMA, which guides the model to better distinguish important regions between the source and target domains.As shown in Fig \ref{fig:cyclegan} we incorporate an attention mechanism in both the decoder and encoder to generate feature maps that highlight important information. This has led to significant improvements in the image generation quality. The attention map is generated by the auxiliary classifier and embedded into the generator and discriminator, respectively. The CAM in the generator allows the model to perceive the significant differences between the two domains (A, B), while the CAM in the discriminator helps distinguish between real and fake samples and enables the gradient to generate better samples.
$\eta$, where $\eta_s(x)$ represents the probability that x comes from $X_S$.$\eta_{D_t}$ and $D_t(x)$  are trained to discriminate whether x comes from $X_t$ or $G_{s \rightarrow t}\left(X_s\right)$.
\begin{equation}
\begin{aligned}
L_{c a m}^{s \rightarrow t}=-\left({E}_{x \sim X_s}\left[\log \left(\eta_s(x)\right)\right]+{E}_{x \sim X_t}\left[\log \left(1-\eta_s(x)\right)\right]\right)
\end{aligned}
\end{equation}
\begin{equation}
\begin{aligned}
L_{\text {cam }}^{D_t}={E}_{x \sim X_t}\left[\left(\eta_{D_t}(x)\right)^2\right]+{E}_{x \sim X_s}\left[\left(1-\eta_{D_t}\left(G_{s \rightarrow t}(x)\right)^2\right] .\right.
\end{aligned}
\end{equation}
Finally, we joint the encoders, decoders,discriminators, and auxiliary classifiers as our final Loss:
\begin{equation}
\min _{G_{s \rightarrow t}, G_{t \rightarrow s}, \eta_s, \eta_t} \max _{D_s, D_t, \eta_{D_s}, \eta_{D_t}} \lambda_1 L_{l s g a n}+\lambda_2 L_{\text {cycle }}+\lambda_3 L_{\text {identity }}+\lambda_4 L_{\text {cam }},
\end{equation}

Among them $\lambda_1$=1, $\lambda_2$=10, $\lambda_3$=100,$\lambda_4$=1000,
$\mathrm{~L}_{\text {lsgan }}=\mathrm{L}_{\text {lsgan }}^{\mathrm{s} \rightarrow \mathrm{t}}+\mathrm{L}_{\text {lsgan }}^{\mathrm{t} \rightarrow \mathrm{s}}$,$\mathrm{L}_{\text {cycle }}=\mathrm{L}_{\text {cycle }}^{\mathrm{s} \rightarrow \mathrm{t}}+\mathrm{L}_{\text {cycle }}^{\mathrm{t} \rightarrow \mathrm{s}}$,
$\mathrm{L}_{\text {identity }}=\mathrm{L}_{\text {dentity }}^{\mathrm{s} \rightarrow \mathrm{t}}+ 
 \mathrm{L}_{\text {identity }}^{\mathrm{t} \rightarrow \mathrm{s}}$,
$\mathrm{L}_{\text {cam }}=\mathrm{L}_{\text {cam }}^{\mathrm{s} \rightarrow \mathrm{t}}+\mathrm{L}_{\text {cam }}^{\mathrm{t} \rightarrow \mathrm{s}}$

We put the CAM attention module at the end of the encoder in the generator of CycleGAN, which helps the generator to better capture the key information in the deep features in the image at the time of image transformation. Specifically, it uses global average pooling to calculate the average of each channel and uses it as a measure of channel importance. Then, by weighting the channels, the channels with higher importance are taken into account more, which enhances the expressiveness and generalization ability of the generator.

\section{Experiment}
\subsection{Dataset}
We utilized the SODA10M dataset\citep{han2021soda10m} as our primary training dataset. SODA10M is a high-quality driving scene dataset, collected via crowdsourcing methods, with images at 1080P+ resolution. The dataset consists of two parts, one with ten million labeled images and another with twenty thousand unlabeled images, categorized into six primary car scene categories including pedestrian, cyclist, car, truck, tram and tricycle. We chose the car category as our primary training set, which only contained images of urban scenes on sunny days in Shanghai. The  test set contained images from a variety of different scenes,as shown in Tab \ref{tab:soda10m}.The training dataset consists of a single scene for each city, location, weather, and time period, whereas the test set contains different categories for each of these criteria. Notably, there are 1656 scenes that take place at night in the test set.

\begin{table}[ht]\small
\caption{\label{tab:soda10m}SODA 10M in detail.\\}
\centering\resizebox{\textwidth}{!}{
\begin{tabular}{l|ccc|ccc|ccc|cc}
\toprule
 & & City&  & &Location& & &Weather && Period&\\
&Shanghai & Shenzhen& Guangzhou & Citystreet &Highway& Countryroad& Clear&Overcast &Rainy& Daytime&Night\\
\midrule
Train  &5000 & 0& 0 & 5000 &0& 0& 5000&0 &0& 5000&0\\
Test &2596 & 1435& 969 & 1587 &2744& 674& 2023&2328 &649& 3344&1656\\
\bottomrule
\end{tabular}}
\end{table}

We used 5000 labeled training images for training CDMA, which only contained scene vehicles on sunny days during the day and city street scenes. Additionally, we selected 5000 labeled test images, which included day and night, clear, overcast, and rainy weather, as well as city street, highway, and country road scenes. The model was not exposed to the test set content prior to testing. For image generation, we chose 1000 high-quality photos with dark scenes from the test set, and conducted generative adversarial learning with 1000 daytime photos from the training set to train a model with dark scene characteristics.

\subsection{Data augmentation}
To enhance the original limited source domain dataset, we added Gaussian noise, random mask, and random brightness, contrast, and saturation adjustments. Adding Gaussian noise can simulate image noise in real-world scenarios, making the model better suited to handle real-world images. Adding random masks can simulate occlusion, partial target missing, and other complex scenarios, making the model better adapted to object detection tasks in complex scenes. Random adjustments to brightness, contrast, and saturation can increase the diversity of data and improve the model's robustness by adjusting the image's brightness, contrast, and saturation. Random brightness adjustment can simulate images under different lighting conditions, allowing the model to better adapt to real-world scenarios. Contrast adjustment can increase the contrast of the image, making the target more prominent and easier to detect. Saturation adjustment can increase or decrease the color saturation of the image, allowing the model to better adapt to different color distributions. These data augmentation methods can improve the model's generalization ability, prevent overfitting, and help the model better adapt to different scenarios and data distributions.

Using these three methods of data augmentation, we increased the original training set source domain images from 5000 to 20000. Each operation added an additional 5000 training samples to the dataset.
\subsection{Evaluation criteria}
According to the previous work, we selected four evaluation indicators, Recall, Precison, F1-Score, and mAP. Recall is a measure of the sensitivity of CDMA, which is how many examples of all the cars CDMA predicted the model was able to correctly identify. precison refers to the total proportion of the actual labels among all the samples predicted as positive examples by our model, which is used to measure the accuracy of the model. F1-Score is a measure of model performance, which is the harmonic mean of precision and recall, and can comprehensively evaluate the classification performance of CDMA. mAP synthesizes the AP values of multiple object categories and can be used to comprehensively evaluate the performance of object detection models.
\begin{equation}
\begin{aligned}
\text { Precision }=\mathrm{TP} /(\mathrm{TP}+\mathrm{FP})
\end{aligned}
\end{equation}
\begin{equation}
\begin{aligned}
\text { Recall }=\mathrm{TP} /(\mathrm{TP}+\mathrm{FN}) \\
\end{aligned}
\end{equation}
\begin{equation}
\begin{aligned}
F_1=2 \cdot \frac{\text { precision } \cdot \text { recall }}{\text { precision }+ \text { recall }}
\end{aligned}
\end{equation}
\subsection{Baselines}
In the study, we compared CDMA with three baseline models. Faster R-CNN\citep{Ren2015faster} utilizes RPN (Region Proposal Network) to generate candidate regions and the ROI Pooling layer to extract features. RetinaNet\citep{lin2017focal} uses Focal Loss to solve the class imbalance problem in object detection, and FPN to extract multi-scale features. Focal Loss is a loss function that can make the model pay more attention to samples that are difficult to classify, thereby improving the accuracy of the model. The RetinaNet algorithm has a good detection effect on small targets, a fast training speed, and relatively high accuracy. Additionally, it can also be used for tasks such as instance segmentation.YOLOv5\citep{zhu2021yolov5} is characterized by its fast speed, relatively high accuracy, and its ability to perform real-time object detection. YOLOv5 transforms the object detection task into a regression problem by using the idea of the YOLO (You Only Look Once) algorithm. The algorithm divides the entire image into multiple grids, with each grid responsible for detecting objects within it. Subsequently, a neural network directly outputs the location and category information of the target. YOLOv5 also employs techniques such as multi-scale prediction and adaptive training to improve the detection effect. YoloV7\citep{wang2023yolov7} is a state-of-the-art object detection model that predicts bounding boxes and class probabilities directly from input images. It uses a fully convolutional architecture and a modified DarkNet backbone to improve accuracy. Additionally, it employs spatial pyramid pooling and feature fusion to better capture object context and improve detection performance. YoloV8 is the latest version of the Yolo object detection model family, featuring new features and optimizations. It introduces a hybrid backbone with ResNet and DarkNet layers for better feature extraction and accuracy. YoloV8 also employs a novel attention mechanism that selectively weights feature maps to improve detection performance and reduce computational overhead. 

In the domain adaptation field, DA Faster R-CNN \citep{chen2018domain}incorporates domain adaptation techniques into Faster R-CNN. It utilizes a method called Self-Adversarial Network to minimize the differences between the source and target domains, thereby improving the performance of object detection. Specifically, a feature layer was used to extract features from the source domain and the target domain, and a domain classifier was used to distinguish the source domain from the target domain. At the same time, the model also uses an inverse gradient layer to combat the difference between the source domain and the target domain, so as to achieve domain adaptation. Similarly, Domain Adaptive Faster R-CNN is based on Faster R-CNN and has better scalability. This means that the performance can be further improved by increasing the depth of the network, adding more feature layers, etc. However, due to the high complexity of domain adaptation training, the training time of this architecture is relatively long.

Another approach, known as Cross-Domain Adaptive Teacher for Object Detection (CDAT)\citep{liAT2022}, employs a cross-domain adaptive teacher model for training. The goal is to address the domain gap between a domain with annotations (source) and a domain of interest without annotations (target). The core idea of this algorithm is to establish a bridge between the source and target domains in object detection. By training a teacher model in the source domain, it is then applied to the object detection task in the target domain. The teacher-student framework has been effective in semi-supervised learning, but it can generate low-quality pseudo labels due to domain shift.  To mitigate this problem, the CDAT model leverages domain adversarial learning and weak-strong data augmentation.  Feature-level adversarial training is used in the student model to ensure domain-invariant features.  Weak-strong augmentation and mutual learning between the teacher and student models are also employed to enable the teacher model to learn from the student model without being biased to the source domain. Feature-level adversarial training in the student model ensures domain-invariant features, which are essential for cross-domain object detection. Weak and strong data augmentation and mutual learning between teacher and student models enable the teacher model to learn knowledge from the student model without bias to the source domain.

\subsection{Experiment details}
Due to GPU limitations, our target domain generation architecture takes an input image with a resolution of 256×256 pixels and produces an output image with the same resolution. To minimize image degradation during the downsizing and upsizing process, we resize the 1920×1080 image to 1080×1080 after using the Laplacian pyramid, which is created using high-resolution blurred images\citep{engin2018cycle}. To obtain the high-resolution image, we replace the top layer of the Laplacian pyramid with the low-resolution image after image generation, and carry out the Laplacian upgrading process as usual. The image is then converted back to 1080×1080 and resized to 1920×1080 again, as shown in the figure.

Table \ref{tab:hyperparameters}outlines the hyperparameters of our model, which we trained using two TITAN xp under the PyTorch-based framework.

\begin{table}[h]\small
\caption{The hyperparameters of our model.\label{tab:hyperparameters}}
\centering
\begin{tabular}{l|c|c}
\hline \text { Models } & \text { Parameters } & \text { Value } \\
\hline \multirow{5}{*}{\text {Image generation}} & \text { Batch Size } & 1 \\
& \text { Image size } & 256*256 \\
& \text { Learning rate } & 0.0001 \\
& \text { Iterations } & 1000000 \\
& \text { Optimizer } & Adam \\
\hline \multirow{6}{*}{\text { Object detection }} & \text { Batch Size } & 8 \\
& \text { Anchors size } & [8,16,32] \\
& \text { Epochs } & 30 \\
& \text { Learning Rate } & 0.001  \\
& \text { Optimizer } & Adam \\
& \text { Momentum } & 0.9 \\
\bottomrule
\end{tabular}
\end{table}

\section{Result}
\subsection{Domain adaptation detection result}

\begin{table}[ht]\small
\caption{Domain adaptation detection result.\label{tab:Domain adaptation detection result}}
\centering
\begin{tabular}{lcccc}
\toprule
& F1-Score & Recall & Precison & mAP\\
\midrule
Attention RCNN  & 0.3757 & 60.12$\%$	&	27.32$\%$	&	41.42$\%$	\\
\hspace{1em} + Cyclegan &  0.4573	& 73.26$\%$	&	33.24$\%$	&	54.92$\%$	 \\
\hspace{1em} + Data augmentation & 0.4383	& 73.37$\%$	&	31.25$\%$	&	57.69$\%$		 \\
\hspace{1em} + Attention mechanism in cyclegan & 0.4891	& 73.88$\%$	&	36.55$\%$	&	59.97$\%$		 \\
\bottomrule
\end{tabular}
\end{table}

\begin{figure}[ht]
\centering
\includegraphics[width=1\linewidth]{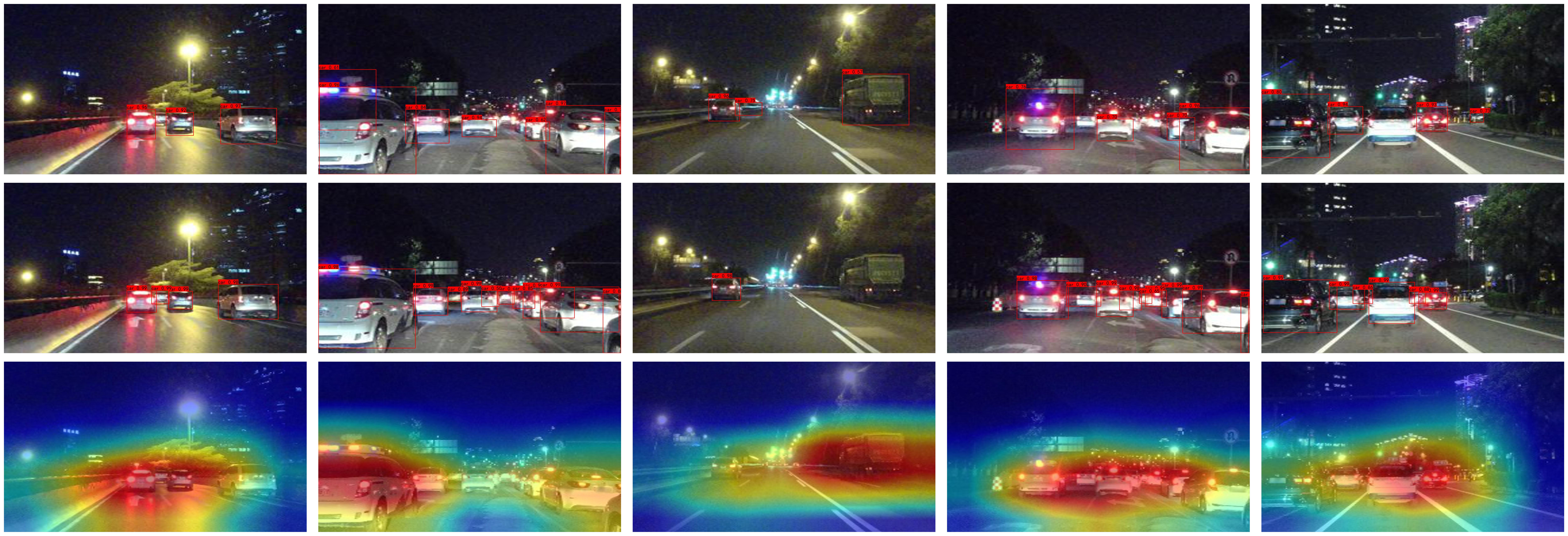}
\caption{\label{fig:Domain adaptation detection result}Domain adaptation result.The image consists of three rows. The first row displays the detection results obtained using the basic Faster R-CNN model. The second row showcases the detection results obtained after utilizing our training strategy and the improved algorithm. The third and final row displays the attention heat map of the object detector.}
\end{figure}

Although our model has shown improvement compared to mainstream object detection algorithms, the improvement is not as significant for object detection in dark scenes. Therefore, we implemented a domain adaptive training strategy for the model, as shown in Section 4. The training results are presented in Table \ref{tab:Domain adaptation detection result}. After fine-tuning with images generated by the adversarial training network, the model demonstrated significant improvement. Specifically, using only dark scene images generated by CycleGAN resulted in a 0.06 and 13.14$\%$ increase in F1-Score, Recall, Precision, and mAP, respectively. After implementing data augmentation and introducing the CAM attention mechanism in CycleGAN, the F1-Score, Recall, Precision, and mAP were increased by 0.1, 13.76$\%$, 9.23$\%$, and 18.55$\%$, respectively, compared to the original model test results.We can see the results more clearly in Fig \ref{fig:Domain adaptation detection result}.

The first column displays the basic Faster RCNN detection results, the second column shows the output from our model, and the third column visualizes the attention mechanism of Faster RCNN. It is apparent that the basic Faster RCNN has a relatively poor detection effect. Not only are the label box scores low, but also close and distant cars are often missed, and even large cars are mislabeled as "CAR". After implementing our cross-domain training strategy and making improvements, the detection effect significantly improved. Our model can now detect cars at long distances and in dark environments, generating appropriate detection boxes.
\begin{table}[ht]\small
\caption{Contrast experiment on SODA 10M.\label{tab:Contrast experiment on SODA 10M}}
\centering
\begin{tabular}{lc@{~}cc@{~}ccc@{~}c}
\toprule
& F1-Score && Recall && Precison & mAP&\\
\midrule
Faster R-CNN\citep{Ren2015faster} & 0.3567	&& 57.66$\%$	&& 25.83$\%$	& 39.29$\%$ \\
RetinaNet\citep{lin2017focal} & 0.4673	&& 53.43$\%$	&&	40.74$\%$	&	40.16$\%$  \\
Yolov5\citep{zhu2021yolov5}  & 0.4794	&& 48.02$\%$	&&	47.86$\%$&	40.07$\%$    \\
Yolov7\citep{wang2023yolov7}  & 0.5641	&& 51.52$\%$	&&	62.31$\%$	&	50.82$\%$    \\
DA-Faster RCNN\citep{chen2018domain}  & 0.4872	&& 62.47$\%$	&&	28.92$\%$	&	43.68$\%$    \\
Yolov8  & \textbf{0.5874}	&& 53.35$\%$	&&	\textbf{64.76$\%$}	&	52.65$\%$    \\
CDAT \citep{liAT2022}& 0.5486	&&  69.17 $\%$	&&	34.29$\%$	&	55.73$\%$   \\
CDMA & 0.4891	&& \textbf{73.88$\%$} &($\uparrow$16.22$\%$)	&	36.55$\%$ 	&	\textbf{59.97$\%$} &($\uparrow$7.32$\%$)	\\
\bottomrule
\end{tabular}
\end{table}

According to the Table \ref{tab:Contrast experiment on SODA 10M}, our model saw significant improvements during training with SODA 10M. Compared to the original RestiNet, YOLOv5, our cross-domain fine-tuned model showes notable gains. Specifically, F1-Score and mAP increases by 0.0218 and 19.81$\%$, 0.097 and 19.90$\%$, respectively. While our model has lower precision rates than RestiNet and YOLOv5 by 4.25$\%$ and 11.31$\%$, the recall of our model improves significantly by 20.45$\%$ and 25.86$\%$. Compared to yolov7 and yolov8,Recall and mAP increases by 22.36$\%$ and 9.15$\%$,20.53$\%$ and 7.32$\%$.Overall, this is a promising outcome. Although there's room to improve the accuracy of our model for car identification and make more precise detection box, it performs better in detecting vehicles in the picture, which is the primary goal of our car detection scenario.

\subsection{Target domain image generation results}
We handpicked 1000 high-quality photos of dark scenes from the test set in SODA10M, as there were several uninformative photos present. We then employed generative adversarial learning with 1000 daytime photos from the training set to train the model with the characteristics of dark scenes.

Although CycleGAN can simulate dark environments well, it often fails to handle the crucial details of dark vehicles with care. As depicted below, night lights appear to be a characteristic of the scene rather than one of dark vehicles in CycleGAN's understanding, despite simulating the dark scene well. However, for vehicular objects, it simply reduces the overall brightness. Consequently, the added lights on the car cannot be placed in their appropriate positions (the headlights and the rear lights) which has a certain misleading effect on our object detector model's training.

\begin{figure}[ht]
\centering
\includegraphics[width=1\linewidth]{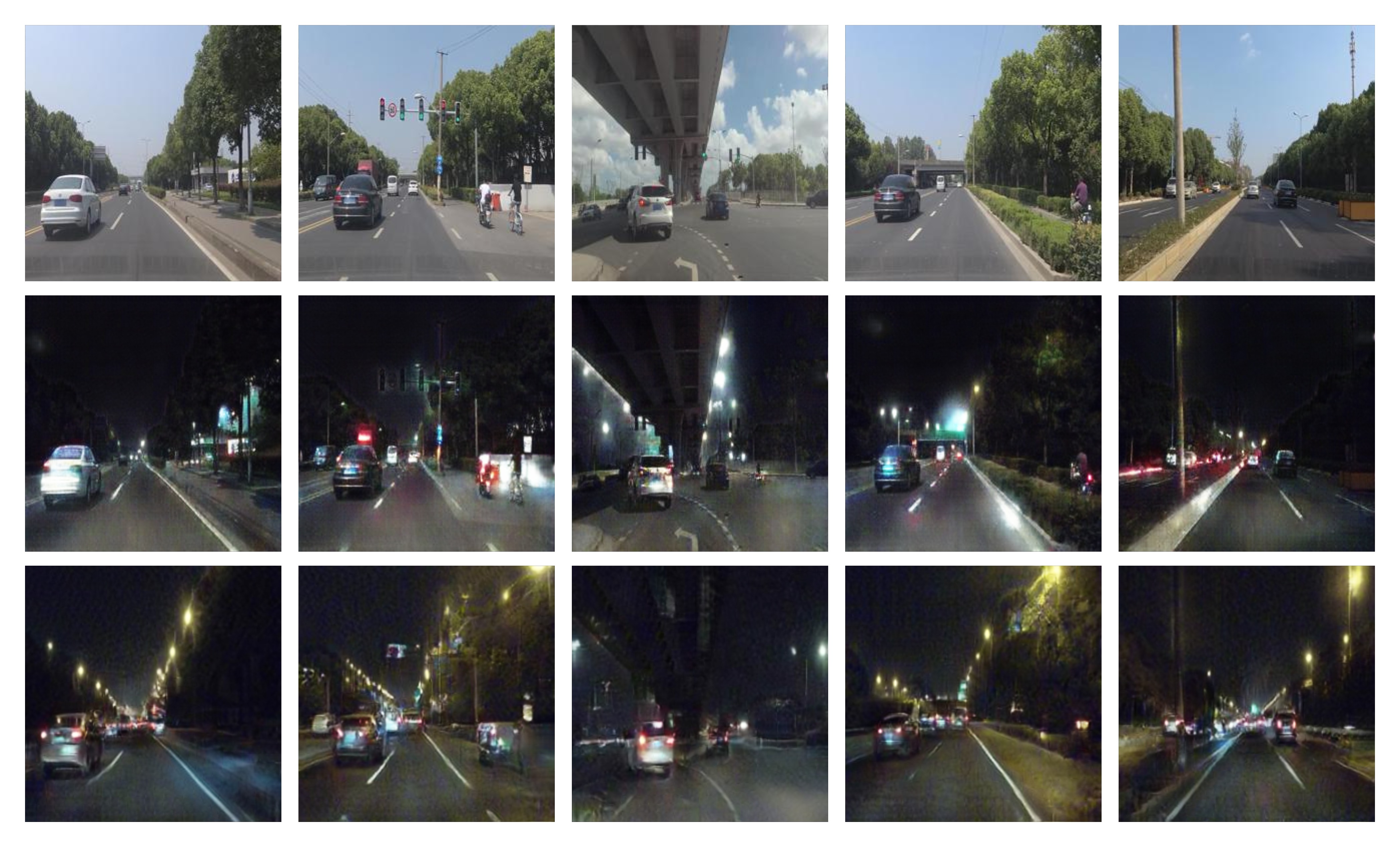}
\caption{\label{fig:resultcyclegan}Comparison of CycleGAN with and without Attention Mechanism.The first row displays the source domain image, followed by the second row which showcases the target domain image generated by CycleGAN. The third row displays the image generated by CycleGAN with the attention mechanism.}
\end{figure}

\begin{figure}[ht]
\centering
\includegraphics[width=0.9\linewidth]{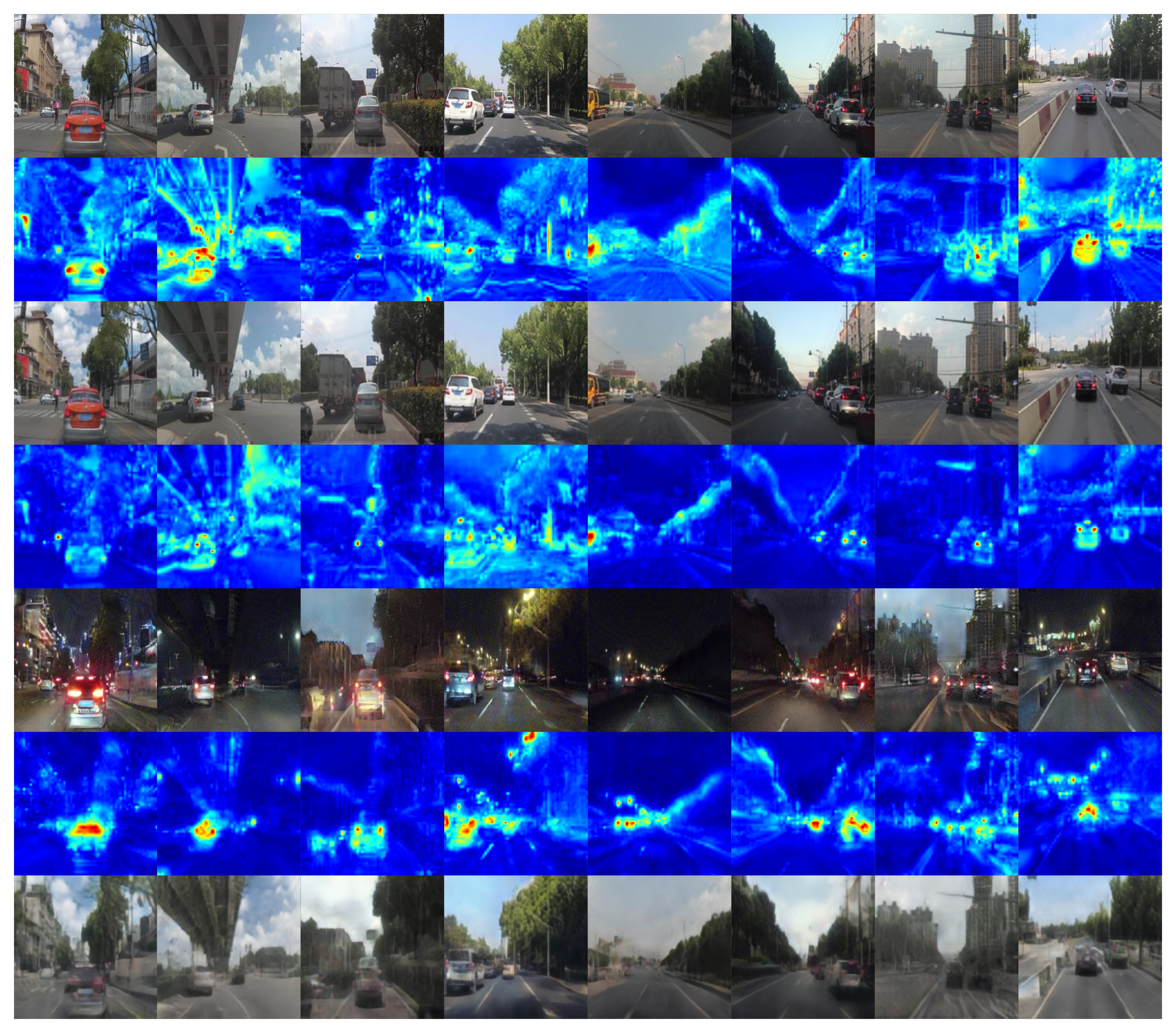}
\caption{\label{fig:image attention}Target domain image generation result. The first row displays the source domain image, followed by the second row which showcases the heat map generated by the source domain generator. The third row displays the source domain image generated by the generator. The fourth and fifth rows show the heat map and image generated by the target domain generator, respectively. The sixth row displays the heat map of the source domain picture generated by the source domain generator using the target domain generated picture, while the seventh row displays the source domain picture generated by the target domain generated picture.}
\end{figure}

With the help of the attention mechanism, generative adversarial training has started to focus more on the primary difference between day and night vehicles during model training. In the region near the headlights, the attention heat map appears redder than other areas, and the model is now able to detect the changes in the key parts of the car while adapting to the overall style changes,as shown in Fig \ref{fig:resultcyclegan}.
As shown in Fig \ref{fig:image attention},by incorporating the attention mechanism, generative adversarial training began to focus more on the primary differences between day and night vehicles during model training. In the area near the headlights, the attention heat map appeared more red than in other areas, indicating that the model was paying closer attention to changes in key parts of the car while the overall style was also changing.

\subsection{Object detection results}
\begin{table}[ht]\small
\caption{Object detection results.\label{tab:Object detection results}}
\centering
\begin{tabular}{lcccc}
\toprule
& F1-Score & Recall & Precison & mAP\\
\midrule
Faster R-CNN  & 0.3567	& 57.66$\%$	& 25.83$\%$	& 39.29$\%$ \\
Attention R-CNN &  0.3757	& 60.12$\%$	&	27.32$\%$	&	41.42$\%$	 \\
\hspace{1em} w/o attention mechanism in the head & 0.3670	& 59.40$\%$	&	26.55$\%$	&	40.78$\%$	 \\
\hspace{1em} w/o attention mechanism in the tail & 0.3558	& 59.93$\%$	&	25.30$\%$	&	41.15$\%$	 \\
\hspace{1em} w/o attention mechanism & 0.3619	& 58.81$\%$	&	26.14$\%$	&	40.71$\%$	 \\
\bottomrule
\end{tabular}
\end{table}

As shown in Table \ref{tab:Object detection results}, we utilized the following methods for our dataset: 1) Faster R-CNN 2) Attention R-CNN  3) Attention-RCNN without attention mechanism in the head4) Attention-RCNN without attention mechanism in the head 5) Attention-RCNN without attention mechanism . By only training with the source domain training set, our model (Attention R-CNN)achieved an F1-Score, Recall, Precision, and mAP that were 0.02, 2.46$\%$, 1.49$\%$, and 2.13$\%$ higher than Faster-RCNN, respectively.

\subsection{Ablation experiment}
To demonstrate the effectiveness of our detector module, we conducted ablation experiments by removing the head attention mechanism, tail attention mechanism, and GIOU loss function, respectively. The results indicate that compared to the tail attention mechanism, the head attention mechanism has a stronger impact on model performance, and the test results significantly dropped after removing the head attention mechanism. Additionally, the use of GIOU and the application of the attention mechanism have a similar positive impact on the model performance.

For our training strategy, ablation experiments were conducted on the five different object detectors (1, 2, 3, 4, 5) presented in Table 1. These experiments included 1) training in the source domain, 2) fine-tuning, 3) fine-tuning for image generation and data augmentation based on CycleGAN, and 4) attention-based CycleGAN and data-augmented fine-tuning. Regardless of the object detector used, there was a significant improvement from training solely in the source domain to fine-tuning with images generated by the generative adversarial network for domain adaptation.
\begin{table}[h]\small
\caption{Ablation experime.\label{tab: Ablation experime}}
\centering
\begin{tabular}{llcccc}
\hline \text { Models } & \text { Training Strategy }& F1-Score & Recall & Precison & mAP \\
\hline \multirow{4}{*}{\text {Faster-RCNN}} & \text { Source domain } & 0.3568 &	57.66$\%$ &	25.83$\%$ &	39.29$\%$   \\
& \hspace{1em} + Cyclegan  & 0.4362 &	69.12$\%$ &	31.86$\%$ &	54.41$\%$  \\
& \hspace{1em} + Data augmentati  & 0.4479 &	72.81$\%$ &	32.42$\%$ &	56.78$\%$  \\
& \hspace{1em} + Attention mechanism in cyclegan  & 0.4598 &	73.62$\%$ &	33.43$\%$ &	58.19$\%$  \\

\hline \multirow{4}{*}{ +GIOU} & \text { Source domain } & 0.3619	 & 58.81$\%$ &	26.14$\%$ &	40.71$\%$ \\
& \hspace{1em} + Cyclegan  & 0.4304 &	69.49$\%$ &	31.77$\%$ &	54.64$\%$   \\
& \hspace{1em} + Data augmentati  & 0.4377 &	70.22$\%$ &	31.79$\%$ &	55.68$\%$  \\
& \hspace{1em} + Attention mechanism in cyclegan  & 0.4740 &	74.18$\%$ &	34.83$\%$ &	58.26$\%$  \\

\hline \multirow{3}{*}{+Attention mechanism} & \text { Source domain } & 0.3558 &	59.93$\%$ &	25.30$\%$ &	41.15$\%$  \\
\multirow{3}{*}{in the head}& \hspace{1em} + Cyclegan  &0.4030 &	70.66$\%$ &	32.41$\%$ &	53.28$\%$   \\
& \hspace{1em} + Data augmentati  & 0.4297 &	72.97$\%$ &	30.45$\%$ &	57.42$\%$   \\
& \hspace{1em} + Attention mechanism in cyclegan  & 0.4867 &	\textbf{74.49}$\%$ &	36.14$\%$ &	58.81$\%$  \\
\hline \multirow{3}{*}{+Attention mechanism} & \text { Source domain } & 0.3670 &	59.40$\%$ &	26.55$\%$ &	40.78$\%$ \\
\multirow{3}{*}{in the tail}& \hspace{1em} + Cyclegan  & 0.4404 &	69.71$\%$ &	32.19$\%$ &	54.39$\%$ \\
& \hspace{1em} + Data augmentati  & 0.4172 & 	71.35$\%$ &	29.48$\%$ &	57.11$\%$  \\
& \hspace{1em} + Attention mechanism in cyclegan  & 0.4767 &	73.37$\%$ &	35.30$\%$ &	59.22$\%$   \\
\hline \multirow{4}{*}{+Attention mechanism} & \text { Source domain } & 0.3757 &	60.12$\%$ &	27.32$\%$ &	41.42$\%$  \\
& \hspace{1em} + Cyclegan  & 0.4573 &	73.26$\%$ &	33.24$\%$ &	54.92$\%$  \\
& \hspace{1em} + Data augmentati  & 0.4383 & 	73.37$\%$ &	31.25$\%$ &	57.69$\%$   \\
& \hspace{1em} + Attention mechanism in cyclegan  & \textbf{0.4891} &	73.88$\%$ &	\textbf{36.55}$\%$ &	\textbf{59.97}$\%$ \\

\bottomrule
\end{tabular}
\end{table}

Observing the results of our ablation experiments, we can conclude that each of our components has proven to be useful. The results show that using GIOU instead of the traditional IOU evaluation metric leads to better experimental results. IOU is a commonly used metric for evaluating object detection algorithms, but it has some limitations. For instance, IOU may be inaccurate when evaluating objects of different sizes, shapes, and orientations. To address this issue, GIOU was proposed. In our model, GIOU considers the offset, width difference, and height difference between the predicted box and the ground truth box, making it more sensitive to changes in object shape and orientation. While IOU only considers the overlap area of two boxes when calculating the metric, GIOU introduces a scaling factor by dividing the overlap area by the minimum bounding box area of the two boxes, which better considers changes in object size. GIOU also takes into account the offset between the predicted box and the ground truth box, making it more sensitive to changes in object position. Furthermore, GIOU is continuous and differentiable, making it more stable and better suited for optimizing object detection models.

Regarding the attention ablation experiments with CBAM, we observed that using this module in any way improved model performance to some extent. We suspect that adding this module increased model complexity, leading to better performance. The results also showed that adding CBAM to different positions had varying effects on model performance. In our model, CBAM enhances feature expression by adaptively adjusting the weights of different feature channels through channel attention. This helps improve the model's perception of objects and thus detection accuracy. Additionally, CBAM suppresses background interference by adaptively learning the spatial correlation of different positions in the feature map through spatial attention. This helps reduce false detection rates and improve model robustness. Furthermore, CBAM can adaptively adjust the correlation between different feature channels and spatial positions, thereby improving the model's generalization ability and stability, reducing overfitting, and improving model reliability.

We can verify the superiority of our model framework by determining whether the decision to use Cyclegan had a decisive impact on the experimental results.

\section{Conclusion}

Based on our experiments and results, we can conclude that our proposed cross-domain car detection model is effective in improving target detection accuracy in the target domain. CDMA includes several novelties, such as a complete cross-domain target detection framework, an unpair target domain picture generation module with an integrated convolutional attention mechanism, the use of Generalized Intersection over Union (GIOU) as the loss function of the target detection framework, a target detection model with an integrated two-head Convolutional Block Attention Module, and an effective data enhancement method.

Our experimental results showed that our framework, which incorporates convolutional attention mechanisms designed to enhance the model's focus on the car headlights feature, significantly improved the performance of the cross-domain vehicle target detection model by 40\% compared to the model without our framework.  Furthermore, our approach had a significant impact on cross-domain vehicle recognition.  By fine-tuning the target detector using the image generator in the target domain based on source domain training, we achieved high-accuracy cross-domain target detection even in the absence of the target domain.

Regarding future research directions, our approach can be extended to other domains, such as pedestrian detection or object detection in other scenarios. Future research can explore the effectiveness of our approach in these domains and further improve the accuracy of cross-domain target detection. Additionally, while our approach achieved high-accuracy cross-domain target detection in the absence of the target domain, it still requires labelled source data. Future research can explore the use of unsupervised domain adaptation techniques to further reduce the reliance on labelled data and improve the scalability of our approach. Furthermore, future research can explore the use of other generative models and investigate their effectiveness in cross-domain target detection. Lastly, while our approach achieved significant improvements in cross-domain target detection, there is still room for further improvement. Future research can explore the use of other attention mechanisms or loss functions to further enhance the detection efficiency of the car detector in the target domain.

Overall, our approach demonstrates that it can effectively address the challenges of cross-domain target detection and significantly improve detection accuracy. We believe that our work provides a solid foundation for future research in cross-domain target detection and can inspire new ideas and approaches for improving the accuracy and effectiveness of cross-domain target detection models.

\section*{Declaration}During the preparation of this work we used Chatgpt in order to improve language and readability of our paper. After using this tool/service, all the authors reviewed and edited the content as needed and take full responsibility for the content of the publication. All the authors read and revised the manuscript.





\bibliographystyle{elsarticle-num}
\bibliography{<your-bib-database>}

\end{document}